\documentclass[letterpaper, 10 pt, conference]{ieeeconf}  % Comment this line out if you need a4paper

\IEEEoverridecommandlockouts                              % This command is only needed if 
\usepackage{graphicx}
\usepackage{booktabs}
\usepackage{float}
\usepackage{stfloats}
\DeclareUnicodeCharacter{2212}{-}
\usepackage{amsmath}
\usepackage{booktabs}
\usepackage{multirow}
\usepackage{amssymb}
\usepackage{color}
\usepackage{mathtools}
\usepackage{amsmath}
\usepackage{dsfont}
\usepackage{algorithm}
\usepackage{algpseudocode}
\usepackage{booktabs}   % toprule, midrule, bottomrule 명령어 사용
\usepackage{colortbl}   % 행 색상 변경 (rowcolor)
\usepackage{makecell}   % makecell 명령어 사용시
\usepackage{tabularx}

\title{\LARGE \bf
KiC: Keyword-inspired Cascade for Cost-Efficient \\Text Generation with LLMs
}

\author{Woo-Chan Kim$^{1}$, Ji-Hoon Park$^{1}$, and Seong-Whan Lee$^{1}$
\thanks{*This research was supported by the Institute of Information \& Communications Technology
Planning \& Evaluation (IITP) grant, funded by the Korea government (MSIT) (No. RS-2019-
II190079 (Artificial Intelligence Graduate School Program (Korea University)), and No. IITP-2025-
RS-2024-00436857 (Information Technology Research Center (ITRC)).}
\thanks{$^{1}$W.-C. Kim, J.-H. Park, and S.-W. Lee are with the Department of Artificial Intelligence, Korea University, Anam-dong, Seongbuk-ku, Seoul 02841, Korea.
    {\tt\small \{kimwc620, jhoon\_park, sw.lee\}@korea.ac.kr}}
}

\begin{document}

\maketitle
\thispagestyle{empty}
\pagestyle{empty}

%%%%%%%%%%%%%%%%%%%%%%%%%%%%%%%%%%%%%%%%%%%%%%%%%%%%%%%%%%%%%%%%%%%%%%%%%%%%%%%%
\begin{abstract} Large language models (LLMs) have demonstrated state-of-the-art performance across a wide range of natural language processing tasks. However, high-performing models are typically accessible only via APIs, incurring substantial inference costs. Cascade methods address this by initially employing a cheaper model and escalating to a stronger one only when necessary. Nevertheless, existing cascade approaches struggle to select a reliable representative response and assess the overall reliability of free-form outputs, as they rely on exact text matching. To overcome these limitations, we propose Keyword-inspired Cascade (KiC), a novel framework for cost-efficient free-form text generation. KiC identifies the most representative answer among multiple outputs from a weaker model and evaluates the semantic alignment of other responses with it. Based on the degree of alignment, KiC determines whether to accept the weaker model’s output or escalate to a stronger model. Experiments on three free-form text generation benchmarks show that KiC achieves 97.53\% of GPT-4’s accuracy while reducing API costs by 28.81\% on average, and even outperforms GPT-4 in a specific benchmark.
\end{abstract}


%%%%%%%%%%%%%%%%%%%%%%%%%%%%%%%%%%%%%%%%%%%%%%%%%%%%%%%%%%%%%%%%%%%%%%%%%%%%%%%%code generation~\cite{li2022competition}, and creative text composition~\cite{anil2023palm}
\section{INTRODUCTION} 
Large language models (LLMs) have demonstrated remarkable performance across various domains, including mathematical reasoning, code generation, and creative text composition. Even though recent best-performing LLMs are black-box models with limited internal access, they are widely utilized in academia and industry via API services. However, the performance capabilities of LLMs correspond directly with their computational costs, creating a significant price spectrum across API services~\cite{chen2024frugalgpt}. For instance, GPT-4 incurs approximately \$30 per million input tokens, whereas GPT-3.5-turbo requires only \$0.50 for the same token count, meaning GPT-4 is about 60 times more expensive. The continuous use of the high-performing model APIs imposes a considerable financial burden, which has motivated the development of LLM cascades~\cite{chen2024frugalgpt, varshney2022model}. The cascade methodology initially employs a lower-cost LLM and escalates to a stronger model only when the confidence score indicates low reliability. 

A key challenge in cascade approaches is identifying when the output from the weaker model can be considered sufficiently reliable. Cascade frameworks may incorporate weaker LLMs configured either as white-box models, which provide access to internal representations, or as black-box models~\cite{si2023prompting, xiong2024can}, which reveal only their generated outputs. While white-box models allow for token-level confidence assessment, they require local deployment. In contrast, black-box models can be accessed without local resource constraints and generally demonstrate superior performance, often making them the preferred choice. In such cases, evaluating response reliability must rely solely on the generated outputs without access to internal model states.

\begin{figure}[t]
    \centering
    \includegraphics[width=1\linewidth]{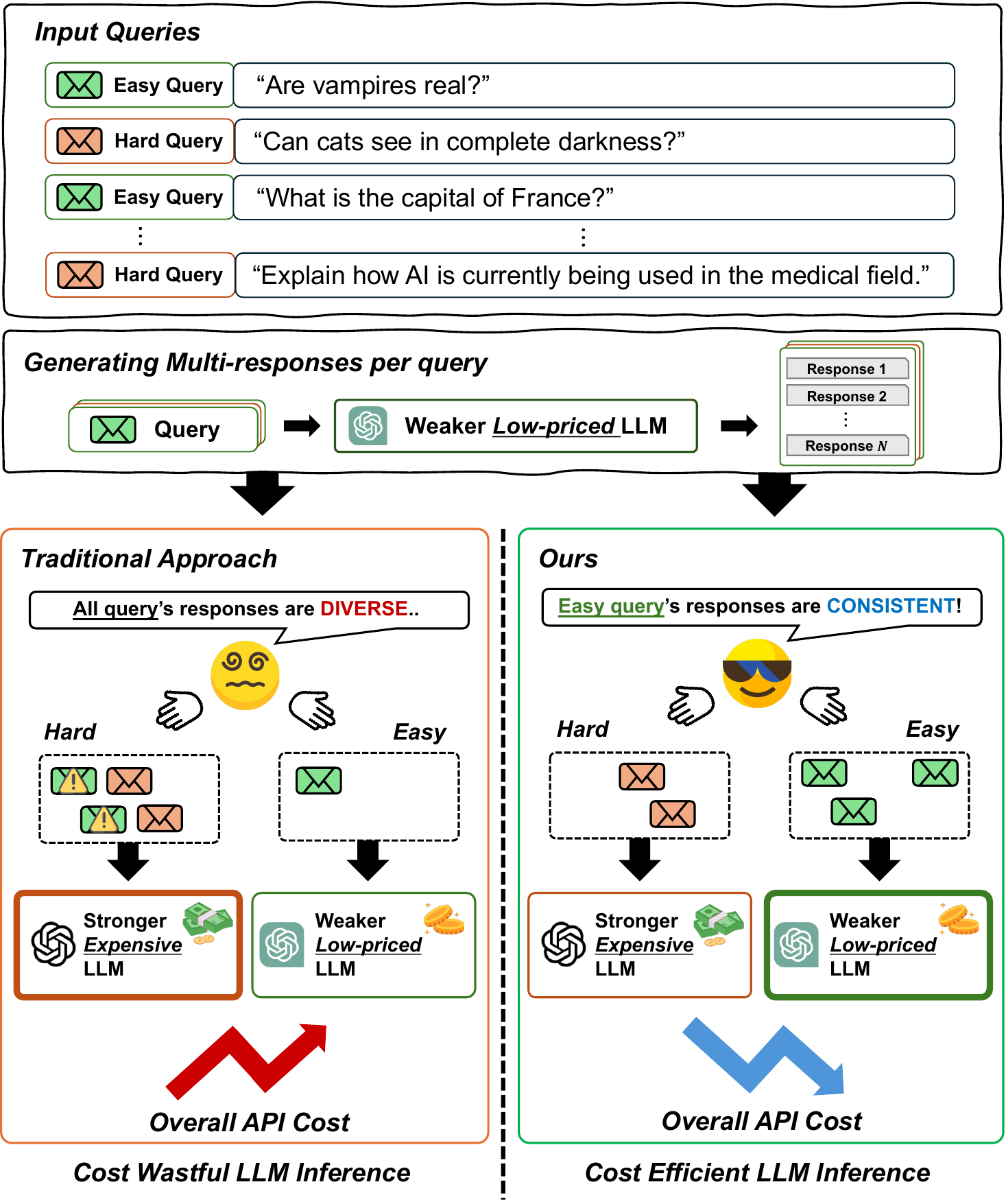}
    \caption{Comparison of cascade strategies between the traditional approach and ours for free-form text generation. Our method identifies easy queries by detecting consistency among responses.}
    \label{fig:1}
\end{figure}

To assess the reliability of weak model outputs, consistency-based evaluation has been proposed as an effective strategy. This method leverages the tendency of LLMs to produce consistent responses when their answers are correct, and diverse responses when they are incorrect~\cite{wang2023self, yue2023large}. This observation motivates the Mixture-of-Thought (MoT) framework~\cite{yue2023large} which integrates Chain-of-Thought (CoT)\cite{wei2022chain} and Program-of-Thought (PoT)\cite{chen2023program}, generating answers through multiple reasoning paths and assessing reliability based on response consistency. While MoT enables cost-efficient cascades for reasoning tasks, it is inherently limited to problems where exact answer matching is possible, such as multiple-choice or numerical questions. Its reliance on the Exact Match (EM) metric limits its use in free-form generation, where semantically equivalent responses may vary in form and response diversity makes majority-vote-based selection of a representative answer unreliable~\cite{chen2024universal}. As a result, EM-based methods struggle not only to judge consistency but also to reliably identify a representative response in free-form generation, highlighting the need for alternative evaluation methods.

\begin{figure*}[t]
    \centering
    \includegraphics[width=1\linewidth]{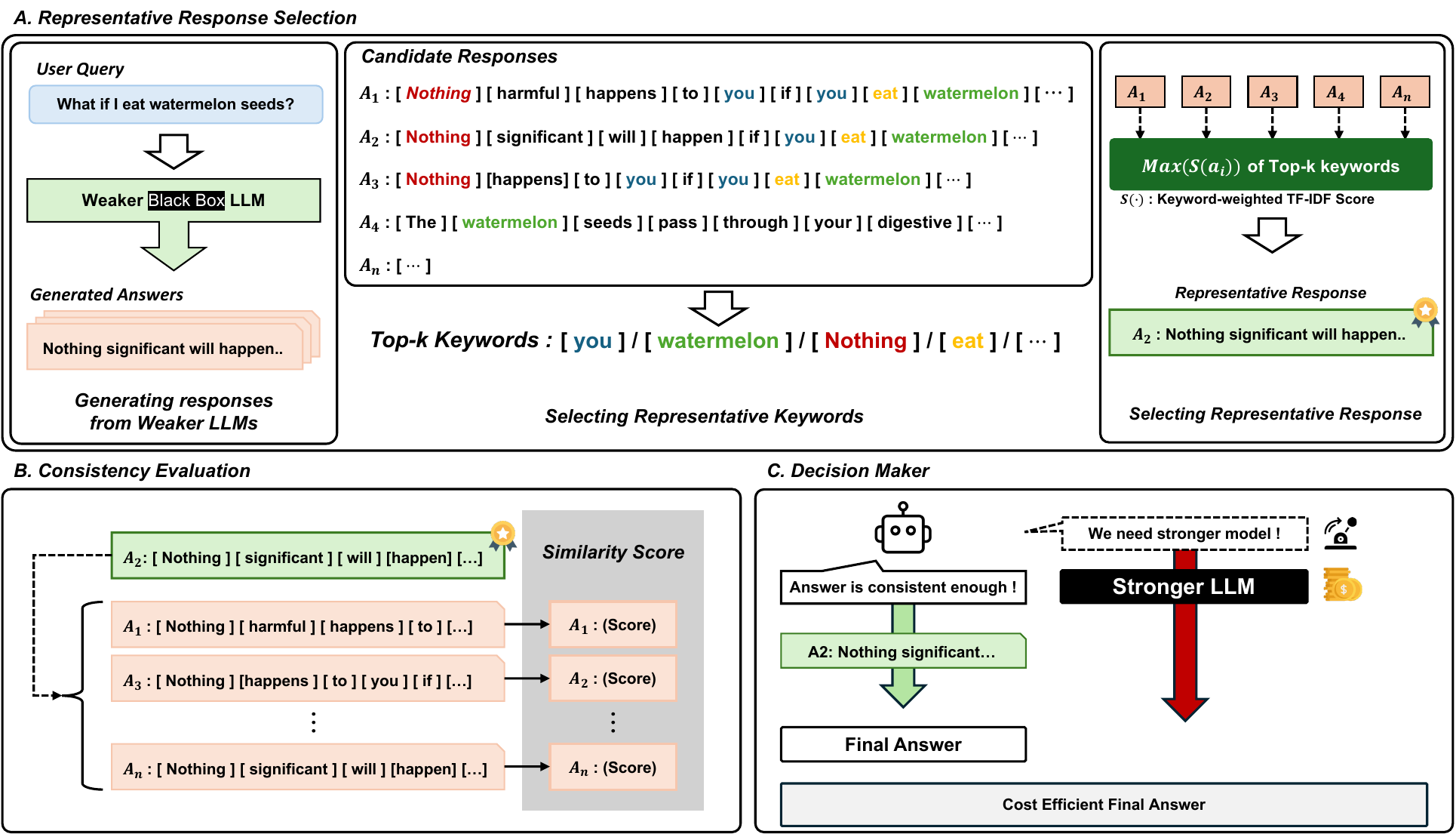}
    \caption{Overview of the Keyword-inspired Cascade (KiC) Framework for Efficient LLM Deployment. This illustrates our approach that combines weaker and stronger LLMs to reduce                             
                                   costs. The framework generates responses from a weaker LLM, selects a representative answer, evaluates consistency, and only escalates to the stronger LLM when necessary.}
    \label{fig:main}
\end{figure*}

In this paper, we propose Keyword-inspired Cascade (KiC), a cascade framework for cost-efficient free-form text generation. To address the limitations of EM evaluation, KiC introduces two core components: a keyword-weighted response selection mechanism and a consistency evaluation mechanism, both aimed at capturing semantic alignment between responses. It first generates multiple candidate responses using a weaker LLM and identifies a representative response based on frequency-weighted keywords and relevance. Then, it computes similarity scores between the representative and each remaining responses using a term importance metric, which assigns higher weights to key terms in the representative response to capture semantic relevance. If enough responses are semantically similar, the representative one is returned; otherwise, a stronger LLM is invoked. By replacing EM with keyword-based semantic comparison, KiC enables cost-efficient cascade decisions in free-form generation, as illustrated in Fig.~\ref{fig:1}.

Extensive experiments across three free-form text generation benchmarks demonstrate that KiC achieves up to 97.53\% of GPT-4's performance, while reducing API costs by an average of 28.81\%. On MMLU-Sociology, KiC outperforms GPT-4 in accuracy while using it only 61\% of the time, demonstrating both cost savings and performance gains.

The contributions of our paper are as follows:
\begin{itemize}
\item We propose KiC, a novel cascade framework with keyword-weighted response selection and consistency evaluation, designed for black-box LLMs in free-form text generation tasks.
\item Our method demonstrates that the cascade model can outperform a single stronger model by leveraging complementary  capbililties.
\item KiC achieves 97.53\% of GPT-4's accuracy while reducing API cost by 28.81\%.
\end{itemize}

\section{RELATED WORKS}

\subsection{LLM Cascades and Cost-Efficient Inference}
The cascade strategy adopts a hierarchical approach in which a lower-capacity, cost-efficient LLM initially generates a response. A more powerful, higher-cost LLM is invoked only when the initial response is deemed insufficient~\cite{lee1990translation, roh2007accurate, lee1990translation}. Early implementations employed fine-tuned verifier models to make routing decisions~\cite{fly2024, prabhakar2020framework}. However, such models often struggled with complex reasoning tasks, where subtle errors could arise even when the generated responses appeared logically sound~\cite{madaan2023selfrefine}. Recent approaches leverage LLMs' tendency to generate consistent answers when confident~\cite{wang2023self, lee2020uncertainty}. The MoT framework~\cite{yue2023large} uses this property to assess response consistency. However, existing cascade methods remain limited in free-form text generation~\cite{lim2000text}, as they primarily rely on EM metrics to assess response consistency~\cite{si2023prompting, yue2023large}. This largely overlooks semantically similar responses with different wording, and prior work largely targets structured outputs over open-domain queries~\cite{kadavath2022language}. In this paper, we address these limitations by proposing a keyword-weighted cascade framework that selects representative responses and evaluates consistency based on term-level similarity rather than exact matching.

\subsection{Text Similarity Assessment in LLMs}
Text similarity assessment has evolved from lexical matching methods (e.g., n-gram, TF-IDF) to embedding-based approaches using distributed~\cite{mikolov2013distributed} and contextual representations. While early methods were computationally efficient, they struggled with semantic variations~\cite{agirre2009study}. Recent advances, such as Sentence-BERT~\cite{reimers-gurevych-2019-sentence} and BERTScore~\cite{Zhang2020BERTScore:}, have significantly improved semantic similarity evaluation. Within LLM cascade frameworks like our KiC approach, however, applying these advanced methods presents challenges—computational overhead from additional model passes~\cite{chen2024frugalgpt}, limited generalization across task domains, and increased latency that undermines cascade efficiency~\cite{yue2023large}. While most existing frameworks rely on exact match or voting-based heuristics, lightweight semantic similarity methods remain underexplored. Our framework enhances lexical similarity with interpretable, keyword-aware weighting, enabling scalable and semantically informed evaluation in cascade settings.

\section{METHODS} In this paper, we propose a cascade framework that efficiently integrates stronger and weaker LLMs for text generation tasks. Fig.~\ref{fig:1} provides an overview of the proposed KiC framework, illustrating the three main stages: representative response selection, consistency evaluation, and decision making. The framework minimizes overall operational costs by invoking large, expensive models only when necessary. While existing cascade approaches have been limited to clear response tasks, our method introduces two key techniques to extend cascades to free-form text generation: (1) a keyword-based representative response selection mechanism, which computes word importance scores to identify the most representative response, and (2) a sentence consistency evaluation mechanism, which compares the representative response with others to identify equivalent responses despite surface-level variations. By combining these techniques, the proposed framework maintains high performance while significantly reducing API usage costs. 

\begin{algorithm}[t]
    \caption{Keyword-inspired Cascade (KiC) Framework}
    \begin{algorithmic}[1]
    \State \textbf{Input:} Query $q$, Weak LLM responses $A = \{a_1, \dots, a_n\}$, Threshold $\tau$
    \State \textbf{Output:} Final response $R_{KiC}$

    \vspace{0.5em}
    \State \textbf{// Step 1: Representative Response Selection}
    \State Extract top-$k$ frequent keywords from $A$
    \For{each $a_i \in A$}
        \State Compute TF-IDF score $S(a_i)$ with keyword weights
    \EndFor
    \State $R^*_{weaker} \gets \arg\max_{a_i} S(a_i)$

    \vspace{0.5em}
    \State \textbf{// Step 2: Consistency Evaluation}
    \State $N_{sim} \gets 0$
    \State $S^* \gets S(R^*_{weaker})$
    \For{each $a_i \in A$}
        \State Compute TF-IDF score $S(a_i)$ using keyword weights from $R^*_{weaker}$
        \If{$S(a_i) \geq S^*$}
            \State $N_{sim} \gets N_{sim} + 1$
        \EndIf
    \EndFor

    \vspace{0.5em}
    \State \textbf{// Step 3: Decision Making}
    \If{$N_{sim} \geq \tau$}
        \State $R_{KiC} \gets R^*_{weaker}$
    \Else
        \State Query stronger LLM for $R_{stronger}$
        \State $R_{KiC} \gets R_{stronger}$
    \EndIf
    \State \Return $R_{KiC}$
    \end{algorithmic}
    \label{alg:kic_updated}
\end{algorithm}

\subsection{Problem Formulation} We formulate our cascade framework as an uncertainty estimation problem for a weaker LLM. Given a query, we first generate multiple responses $A_w = \{a_1, a_2, \ldots, a_n\}$ using the weaker LLM. The core challenge is to assess whether the responses exhibit sufficient consistency to ensure reliability or require escalation to a stronger LLM. To address this challenge, we introduce a three-step decision-making approach. First, we select a representative response using a scoring function based on frequency-weighted TF-IDF. Then, we evaluate the consistency of the remaining responses with respect to the selected representative. Lastly, the representative response is selected as the final output if the number of semantically similar responses exceeds a predefined threshold; otherwise, a stronger LLM is invoked. The overall procedure is summarized in Algorithm 1.

\subsection{Keyword-Weighted Response Selection} We utilize the TF-IDF metric to select a representative response from responses. TF-IDF assigns weights to terms based on their informativeness. It multiplies the Term Frequency (TF), which measures how often a term appears in a given response, by the Inverse Document Frequency (IDF), which reflects the term’s rarity across the entire response set:
\begin{equation}
    \text{TF-IDF}(t, d, D) = \text{TF}(t, d) \times \text{IDF}(t, D),
\end{equation}
where $t$ is a term, $d$ a document (\textit{i.e.}, a response), and $D$ the set of all responses.

While TF-IDF enables the selection of responses with highly informative terms, overemphasizes rare words. This often hinders the identification of responses that are representative of the general response distribution, key aspects for ensuring self-consistency in answer selection. To address this limitation, we extract the top-$k$ most frequent keywords from the response cluster. These keywords receive higher weights during TF-IDF computation, promoting the selection of responses that are both informative and representative. To avoid penalizing short responses, which naturally contain fewer terms, we apply L2-norm normalization to the final score. This ensures that the score reflects the informativeness and representativeness of the response, independent of its length. The final weighted score for each response $a_i$ is defined as: 
\begin{equation} 
    S(a_i) = \frac{1}{\|w_{a_i}\|_2} \sum_{t \in a_i} w_t \times \text{TF-IDF}(t, a_i, A), 
    \label{eq:main}
    \end{equation} where $w_{a_i} = \big[ w_t \times \text{TF-IDF}(t, a_i, A) \big]_{t \in a_i}$ is the vector of weighted term scores for response $a_i$, and $\|w_{a_i}\|_2$ denotes its L2 norm. The keyword weight $w_t$ is defined as: 
    \begin{equation} w_t = \begin{cases} \alpha, & \text{if }t \in \text{top-}k\text{ frequent terms} \\ 1, & \text{otherwise} \end{cases} . 
\end{equation}

After computing scores $S(a_i)$ for each response $a_i\in A_w$, the representative response $R^*_{weaker}$ is determined by selecting the one with the highest score as defined in Eq.~\ref{max}.

\begin{equation}
R^*_{weaker} = \arg\max_{a_i \in A_w} S(a_i).
\label{max}
\end{equation}

\subsection{Consistency Evaluation Framework} As shown in the second module of Fig.~\ref{fig:main}, after selecting the representative response, we assign weights to its keywords and compute frequency-weighted TF-IDF scores for the remaining responses. We assess response consistency using Eq.~\ref{eq:main}. To highlight keyword-level similarity, we assign higher weights to the keywords in the representative response. This ensures that responses with similar term usage receive higher scores. The keyword weight \( w_t \) is defined as:
\begin{equation}
    w_t = 
    \begin{cases}
    \alpha, & \text{if } t \in \text{top-}k \text{ frequent terms} \\
    \beta, & \text{if } t \in \text{keywords in the representative response} \\
    1, & \text{otherwise}
    \end{cases}
    ,
\end{equation}
where the weights are set such that \(1 < \alpha < \beta \), ensuring that keywords appearing in the representative response receive the highest contribution to the final score. Finally, we determine the number of responses, \( N_{\text{sim}} \), whose scores are greater than or equal to the keyword-weighted TF-IDF score of the representative response, denoted as \( S^* \). 

\subsection{Decision Making} This stage corresponds to third stage in Fig.~\ref{fig:main}. Based on whether \( N_{\text{sim}} \) surpasses the decision threshold \( \tau \), the final output \( R_{\text{KiC}} \) is selected as either the representative response from the weaker LLM or the response generated by the stronger LLM, as defined below:

\begin{equation}
    R_{\text{KiC}} = 
    \begin{cases}
    R^*_{weaker}, & \text{if } N_{sim} \geq \tau \\
    R_{stronger}, & \text{otherwise}
    %R_{stronger}, & \text{if } N_{total} < \tau,
\end{cases}
.
\end{equation}

 This decision-making mechanism ensures that we only invoke the more computationally expensive model when the weaker model's responses lack sufficient consistency, thereby optimizing both performance and  API cost efficiency. The optimal value of the threshold $\tau$ is not fixed a priori but is empirically explored and analyzed in the experimental section to ensure a robust balance between accuracy and cost.

\begin{table}[h]
    \centering
    \caption{Performance and Cost Comparison Across Datasets}
    \setlength{\tabcolsep}{2pt} % 열 간 간격 조정
    \renewcommand{\arraystretch}{1.4} % 행 간 간격 조정
    \begin{tabular}{llcccc}
        \toprule
        \textbf{Dataset} & \textbf{Metrics} & \textbf{GPT-3.5-turbo} & \textbf{GPT-4} & \textbf{EM} & \textbf{KiC ($\tau$ = 8)} \\
        \midrule
        \multirow{2}{*}{TruthfulQA} & Accuracy & 52.51 & 69.77 & 67.20 & 63.89 \\
        & Cost (\$) & 0.03 & 1.71 & 1.72 & 1.15 \\
        & GPT-4 Usage & - & 1.00 & 0.89 & 0.26 \\
        \midrule
        \multirow{2}{*}{\shortstack{MMLU-\\Sociology}} & Accuracy & 51.74 & 61.19 & 61.19 & 62.19 \\
        & Cost (\$) & 0.01 & 0.58 & 0.65 & 0.46 \\
        & GPT-4 Usage & - & 1.00 & 0.93 & 0.61 \\
        \midrule
        \multirow{3}{*}{\shortstack{MMLU-\\Professional\\Psychology}} & Accuracy & 47.39 & 54.58 & 54.41 & 54.25 \\
        & Cost (\$) & 0.03 & 1.79 & 1.95 & 1.20 \\
        & GPT-4 Usage & - & 1.00 & 0.91 & 0.48 \\
        \bottomrule
    \end{tabular}
    \label{table:performance-comparison}
\end{table}

\section{EXPERIMENTS}
\subsection{Experimental Settings}

\subsubsection{Datasets}
We evaluate our KiC approach on three free-form text generation datasets:
\begin{itemize}
\item \textbf{TruthfulQA~\cite{lin-etal-2022-truthfulqa}}: A benchmark evaluating the ability of language models to avoid generating misconceptions and produce factually accurate and contextually grounded responses.
\item \textbf{MMLU-Sociology~\cite{hendrycks2021measuring}}: The sociology subset of MMLU, where multiple-choice questions are reformulated as open-ended prompts to assess coherent reasoning and sociological understanding.
\item \textbf{MMLU-Professional Psychology~\cite{hendrycks2021measuring}}: A professional psychology subset of MMLU, rephrased into free-form prompts to evaluate responses grounded in psychological theory, diagnosis, and treatment practices.
\end{itemize}

These datasets are designed for free-form answers and highlight substantial performance differences across model strengths.

\subsubsection{Models and Cascade Configurations}
The cascade architecture consisted of two language models: GPT-3.5-turbo as the weaker LLM and GPT-4 as the stronger LLM. Both models were accessed via publicly available APIs, with fixed decoding parameters to ensure experimental reproducibility. The weaker LLM was configured with a temperature of 1.0, generating 10 responses per query to evaluate response consistency. In contrast, the stronger LLM was set to a temperature of 0.0, producing deterministic outputs used as reference answers.

\subsubsection{Evaluation Metrics}
GPT-4 was employed as an automated evaluator to assess the accuracy on the TruthfulQA, MMLU-Sociology, and MMLU-Professional Psychology datasets. GPT-4 provided binary correctness judgments (\textit{i.e.}, True or False) given the input query, the model-generated answer, and the corresponding reference answer. The reliability of this evaluation method was supported by the alignment between the observed performance gaps of GPT-3.5-turbo and GPT-4 and previously reported differences~\cite{achiam2023gpt}.

\begin{figure*}
    \centering
    \includegraphics[width=1\linewidth]{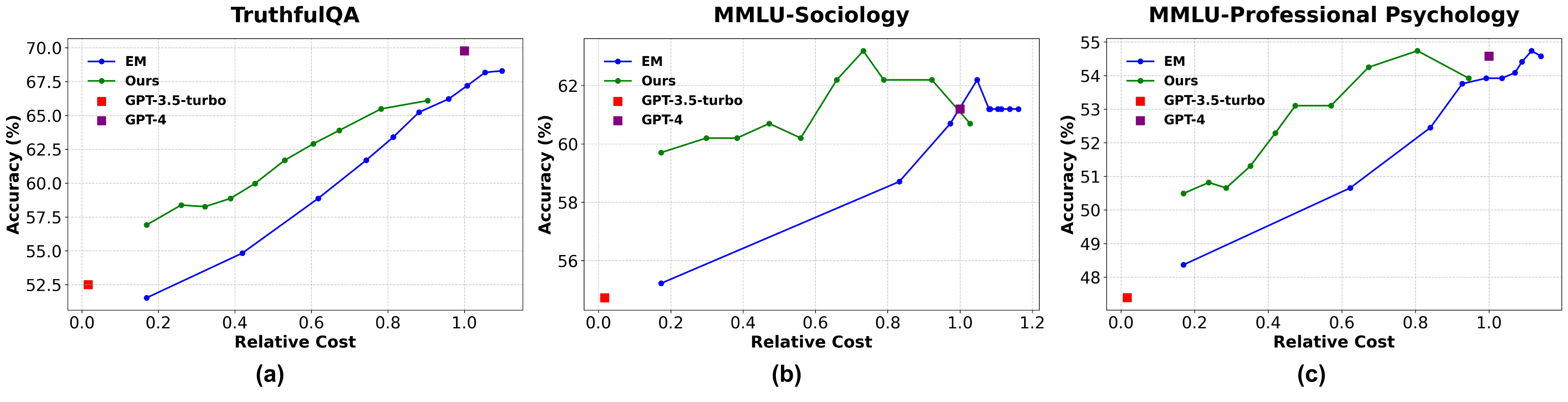}
    \caption{Accuracy vs. Relative cost across three benchmarks. Comparison of our KiC method (green) against EM approach (blue) with GPT-3.5 and GPT-4 baselines as reference points. KiC consistently achieves higher accuracy at comparable costs across all datasets. Each dot represents a different $\tau$ value, increasing sequentially from 1 to 10 from the lower left to the upper right.}
    \label{fig:main}
\end{figure*}

\subsection{Main Results} Fig.~\ref{fig:main} illustrates that KiC achieves substantial cost reductions while maintaining competitive performance across all three datasets, particularly when compared to the EM approach. According to Tab.~\ref{table:performance-comparison}, the EM method achieves performance close to GPT-4 but relies heavily on GPT-4 invocations, resulting in a total cost that exceeds using GPT-4 alone. Despite its high accuracy, this indicates limited practicality due to excessive cost. In contrast, KiC maintains 97.53\% of GPT-4's performance while reducing the GPT-4 invocation rate to 45\%, leading to a 28.81\% overall cost reduction. This corresponds to a 33.61\% cost saving compared to EM at a similar performance level, demonstrating the superior cost-efficiency of our method.
Additionally, as the threshold $\tau$ increases from 1 to 10, KiC consistently exhibits stable and high performance, with the optimal balance between cost and accuracy observed in the range $\tau = 7$ to $9$. This confirms the robustness and reliability of our keyword-based routing strategy across diverse tasks.

\subsubsection{TruthfulQA} As shown in Fig.~\ref{fig:main}(a), the KiC method substantially improves cost-effectiveness compared to the EM baseline on the TruthfulQA benchmark. As shown in Tab.~\ref{table:performance-comparison}, KiC improves accuracy by 11.38 points compared to GPT-3.5-turbo, achieves 91.57\% of GPT-4's performance, and reduces inference cost by approximately 32.75\%. The stable performance-cost curve of KiC highlights its robustness in hallucination-sensitive tasks like TruthfulQA, where LLMs tend to respond consistently to known facts and inconsistently to uncertain ones.

\subsubsection{MMLU Sociology} Fig.~\ref{fig:main}(b) shows that in the MMLU-Sociology benchmark, KiC demonstrated not only strong cost-effectiveness but also the ability to outperform GPT-4 in terms of accuracy. As shown in Tab.~\ref{tab:truthfulqa-transposed}, KiC achieved higher accuracy than GPT-4 when the threshold $\tau$ was set between 6 and 9. This can be attributed to the self-consistency effect observed in the representative answers selected from multiple outputs of GPT-3.5-turbo, which correctly answered instances where GPT-4 failed. Such complementary behavior between the two models led to overall performance improvements. These findings suggest that cascade methods can serve not only as cost-saving mechanisms but also as strategies for enhancing performance through the complementary use of models with different capabilities.

\begin{figure}[t]
    \centering
    \includegraphics[width=0.95\linewidth]{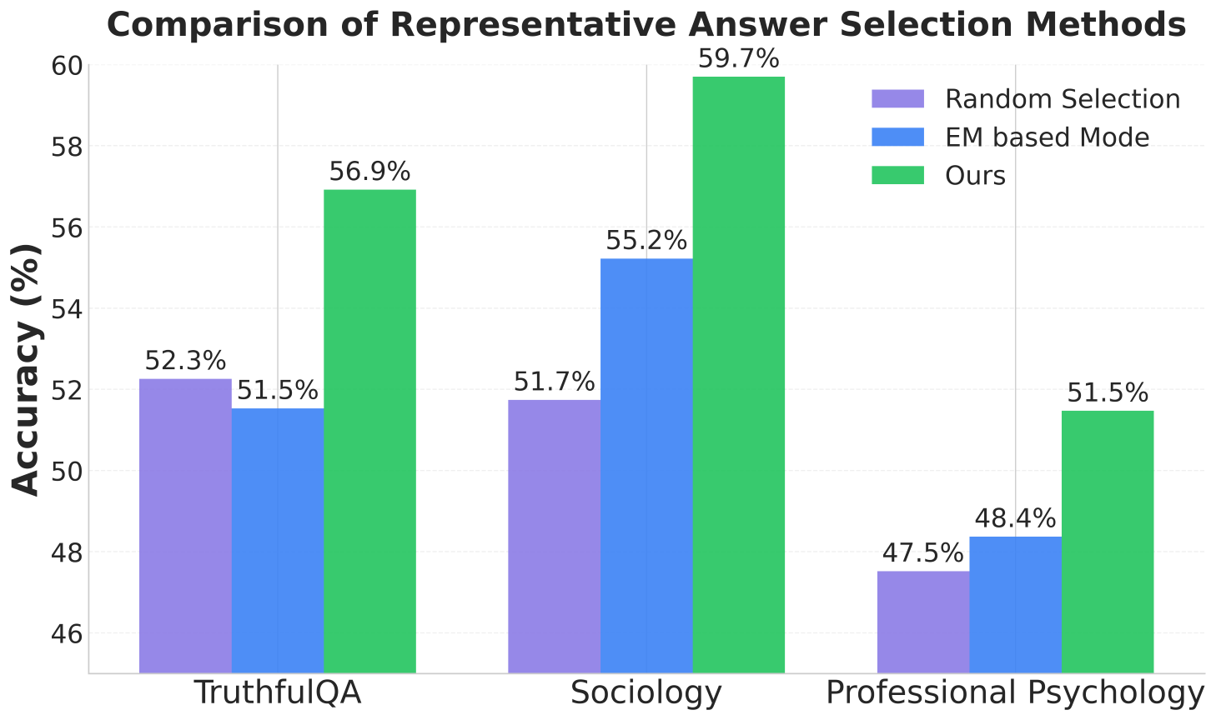}
    \caption{Comparison of different representative response selection methods for free-form text generation on three datasets}
    \label{fig:representative}
\end{figure}

\begin{table}[h] 
    \centering
    \caption{Threshold-wise GPT-4 Usage and Relative Performance (\%) of KiC on MMLU-Sociology}
    \setlength{\tabcolsep}{4pt}
    \renewcommand{\arraystretch}{1.2}
    \begin{tabular}{lcccccc}
        \toprule
        \textbf{Threshold ($\tau$)} & \textbf{5} & \textbf{6} & \textbf{7} & \textbf{8} & \textbf{9} & \textbf{10} \\
        \midrule
        \textbf{Relative Performance (\%)} & 98 & 102 & \textbf{103} & 102 & 102 & 99 \\
        \textbf{GPT-4 Usage (\%)} & 37.3 & 47.8 & 55.7 & 61.2 & 73.6 & 84.6 \\
        \bottomrule
    \end{tabular}
    \label{tab:truthfulqa-transposed}
\end{table}

\subsubsection{MMLU Professional Psychology} As shown in Tab.\ref{table:performance-comparison}, KiC achieves 99.39\% of GPT-4’s performance while reducing GPT-4 usage to 48\%, resulting in a 32.96\% cost reduction. In comparison, the EM method relies on GPT-4 in 91\% of cases and incurs 108.93\% of GPT-4’s cost, making KiC 38.46\% more cost-efficient at a similar performance level. 
This advantage stems from KiC’s ability to effectively identify core domain-specific terms~\cite{hendrycks2021measuring} such as ``separation-individuation" and ``Thorndike's Law" through its keyword-weighted scoring mechanism. By assigning higher weights to keywords during TF-IDF-based similarity computation, KiC enhances the likelihood of selecting an accurate representative response. These results suggest that KiC is a robust approach even for domain-specific question answering tasks.

\begin{table}[h]
    \centering
    \caption{Accuracy Comparison of Representative Selection Methods Across Benchmarks (\%)}
    \setlength{\tabcolsep}{4pt}
    \renewcommand{\arraystretch}{1.2}
    \small
    \begin{tabular}{lccc}
        \toprule
        \textbf{Approach} & \textbf{TruthfulQA} & \textbf{Sociology} & \textbf{Psychology}\\
        \midrule
        Greedy & 52.51 & 51.4 & 47.39 \\
        Random & 53.27 & 51.74 & 47.52 \\
        EM & 51.53 & 55.22 & 48.37\\
        KiC & \textbf{56.92} &\textbf{59.70} & \textbf{51.47} \\
        \bottomrule
    \end{tabular}
    \label{table:rep_comparison}
\end{table}

\subsection{Representative Selection Analysis}
We evaluated the effectiveness of our keyword-weighted representative selection mechanism by comparing it with several baselines, including greedy selection, random selection, and the conventional EM method. As shown in Fig.~\ref{fig:representative},  our KiC approach consistently outperforms other methods across different benchmarks. On TruthfulQA, as reported in Tab.~\ref{table:rep_comparison}, the EM method achieved lower accuracy (51.53\%) than both random selection (53.27\%) and greedy selection (52.51\%), indicating that traditional EM-based selection is not well suited for free-form text generation tasks. 

In MMLU-Sociology, KiC’s selected representative responses achieved substantially higher accuracy compared to other methods, outperforming the greedy baseline by 8.3\%. This improvement can be attributed to the self-consistency effect~\cite{wang2023self}, where generating multiple responses and selecting the most consistent one enhances overall accuracy. Notably, in MMLU-Sociology, KiC surpassed GPT-4 in performance, as shown in Fig.~\ref{fig:main}, highlighting the significant impact of representative selection on the overall effectiveness of cascade methods.

\section{CONCLUSION}
In this paper, we proposed the KiC framework, a cost-efficient approach for LLM inference in free-form text generation. KiC introduces a keyword-weighted representative selection mechanism and a sentence-level consistency evaluation. These techniques address the limitations of exact-match-based cascades and enable effective handling of semantically equivalent but lexically diverse responses. Experiments on three generation benchmarks show that KiC retains around 97.53\% of GPT-4's performance while reducing cost by an average of 28,81\%. On MMLU-Sociology, it even outperformed GPT-4 with 20.68\% cost savings. These results demonstrate KiC's practicality for API-based applications requiring both accuracy and efficiency.

\addtolength{\textheight}{-12cm}   % This command serves to balance the column lengths

%%%%%%%%%%%%%%%%%%%%%%%%%%%%%%%%%%%%%%%%%%%%%%%%%%%%%%%%%%%%%%%%%%%%%%%%%%%%%%%%

%%%%%%%%%%%%%%%%%%%%%%%%%%%%%%%%%%%%%%%%%%%%%%%%%%%%%%%%%%%%%%%%%%%%%%%%%%%%%%%%

%%%%%%%%%%%%%%%%%%%%%%%%%%%%%%%%%%%%%%%%%%%%%%%%%%%%%%%%%%%%%%%%%%%%%%%%%%%%%%%%

% \section*{ACKNOWLEDGMENT}

% The preferred spelling of the word ÒacknowledgmentÓ in America is without an ÒeÓ after the ÒgÓ. Avoid the stilted expression, ÒOne of us (R. B. G.) thanks . . .Ó  Instead, try ÒR. B. G. thanksÓ. Put sponsor acknowledgments in the unnumbered footnote on the first page.

%%%%%%%%%%%%%%%%%%%%%%%%%%%%%%%%%%%%%%%%%%%%%%%%%%%%%%%%%%%%%%%%%%%%%%%%%%%%%%%%

\bibliographystyle{IEEEtran}
\bibliography{REFERENCE}

\end{document}